\documentclass[runningheads]{llncs}

\usepackage{color}
\usepackage{times}
\usepackage{epsfig}
\usepackage{graphicx}
\usepackage{amsmath}
\usepackage{amssymb}

\usepackage{multirow}

\usepackage{graphicx} 

\usepackage{algorithm}
\usepackage{algorithmic}

\usepackage{booktabs}
\usepackage{wrapfig,lipsum,booktabs}
\usepackage{bm}

\newcommand{\eat}[1]{}      

\usepackage{amsmath,amssymb}
\usepackage{amsfonts}
\usepackage{mathtools}
\usepackage{bm} 

\newcommand{\BG}[1]{{\textcolor{blue}{#1}}}

\newcommand{\vct}[1]{\boldsymbol{#1}} 
\newcommand{\mat}[1]{\boldsymbol{#1}} 
\newcommand{\cst}[1]{\mathsf{#1}}  

\newcommand{\vx}{{\vct{x}}}
\newcommand{\vy}{\vct{y}}
\newcommand{\vz}{{\vct{z}}}

\newcommand{\vw}{\vct{w}}

\newcommand{\mU}{\mat{U}}
\newcommand{\mV}{\mat{V}}

\newcommand{\mW}{\mat{W}}

\newcommand{\cN}{\cst{N}}

\newcommand{\cT}{\cst{T}}

\newcommand{\sV}{\mathcal{V}}
\newcommand{\calY}{\mathcal{Y}}
\newcommand{\mL}{\mat{L}}
\newcommand{\mI}{\mat{I}}
\newcommand{\mK}{\mat{K}}

\newcommand{\vf}{\vct{f}}

\newcommand{\R}{\mathbb{R}}

\usepackage{tikz}
\usetikzlibrary{calc,arrows,positioning,shapes}

\usepackage[width=122mm,left=12mm,paperwidth=146mm,height=193mm,top=12mm,paperheight=217mm]{geometry}

\begin{document}


\pagestyle{headings}
\mainmatter

\title{How Local is the Local Diversity? Reinforcing Sequential Determinantal Point Processes with Dynamic Ground Sets for Supervised Video Summarization\vspace{-0.0in}
} 

\titlerunning{Reinforcing SeqDPP with Dynamic Ground Sets for Supervised Video Summarization}

\authorrunning{Yandong Li, Liqiang Wang, Tianbao Yang, Boqing Gong}


\author{Yandong Li$^{1}$  \quad  Liqiang Wang$^{1}$ \quad Tianbao Yang$^{2}$ \quad Boqing Gong$^{3}$ \vspace{-0.00in}
}
\institute{$^1$University of Central Florida, 
$^2$University of Iowa,  
$^3$Tencent AI Lab, USA\\
\email{ lyndon.leeseu@outlook.com,lwang@cs.ucf.edu,\\tianbaoyang@uiowa.edu,boqinggo@outlook.com}
 \vspace{-0.0in}
}

\maketitle
\begin{abstract}
The large volume of video content and high viewing frequency demand automatic video summarization algorithms, of which a key property  is the capability of modeling diversity. If videos are lengthy like hours-long egocentric videos, it is necessary to track the temporal structures of the videos and enforce local diversity. The local diversity refers to that the shots selected from a short time duration are diverse but visually similar shots are allowed to co-exist in the summary if they appear far apart in the video. In this paper, we propose a novel probabilistic model, built upon SeqDPP~\cite{DBLP:conf/nips/GongCGS14}, to dynamically control the time span of a video segment upon which the local diversity is imposed. In particular, we enable SeqDPP to learn to automatically infer \emph{how local the local diversity is supposed to be} from the input video. The resulting model is extremely involved to train by the hallmark maximum likelihood estimation (MLE), which further suffers from the  exposure bias and non-differentiable evaluation metrics. To tackle these problems, we instead devise a reinforcement learning algorithm for training the proposed model. Extensive experiments verify the advantages of our model and the new learning algorithm over MLE-based methods. 
\end{abstract}

\vspace{-0.2in}
\section{Introduction} \label{sIntro}
The Internet age has come to such a new phase that high-definition videos are both ubiquitous and dominant in the IP traffic featured by the boom of video sharing websites, online movies and television shows, and the emerging live video streaming services. Some statistics indicate that about 300 hours of video are uploaded to YouTube per minute and more than 500 million hours of video are watched on YouTube daily. Such a large volume of video content and high viewing frequency demand automatic video summarization algorithms. By distilling important events from the original video and condensing them to a short video clip (or a story board, text description, etc.), video summarization has a great potential in many real-world applications.

Video summarization has been one of the basic research areas in the fields of computer vision and multimedia for decades~\cite{money2008video}. A variety of  techniques have been proposed for different scenarios of video summarization. In general, a good video summary is supposed to describe main events~\cite{hong2009event,khosla2013large,ngo2003automatic} happened in the video and meanwhile remove the video shots that are redundant~\cite{liu2002optimization,zhang1997integrated} and/or unimportant~\cite{lee2012discovering,lu2013story}. 

We consider video summarization as a \emph{diverse} subset selection problem: given a video that can be seen a collection of shots, the goal is to select a subset from the collection to summarize the whole video. This view opens the door for supervised learning approaches to video summarization~\cite{DBLP:conf/nips/GongCGS14,gygli2015video,zhang2016summary,gygli2014creating,DBLP:conf/uai/ChaoGGS15} that fit subset selection models to the video summaries annotated by users. Unlike the conventional unsupervised video summarization methods~\cite{zhang1997integrated,lee2012discovering,ngo2003automatic,lu2013story,hong2009event,khosla2013large,kang2006space,ma2002user}, the supervised ones implicitly infer users' intentions and summarization criteria as opposed to domain experts' handcrafting. 

\begin{figure*}[t]
    \centering
    \includegraphics[width=1 \textwidth]{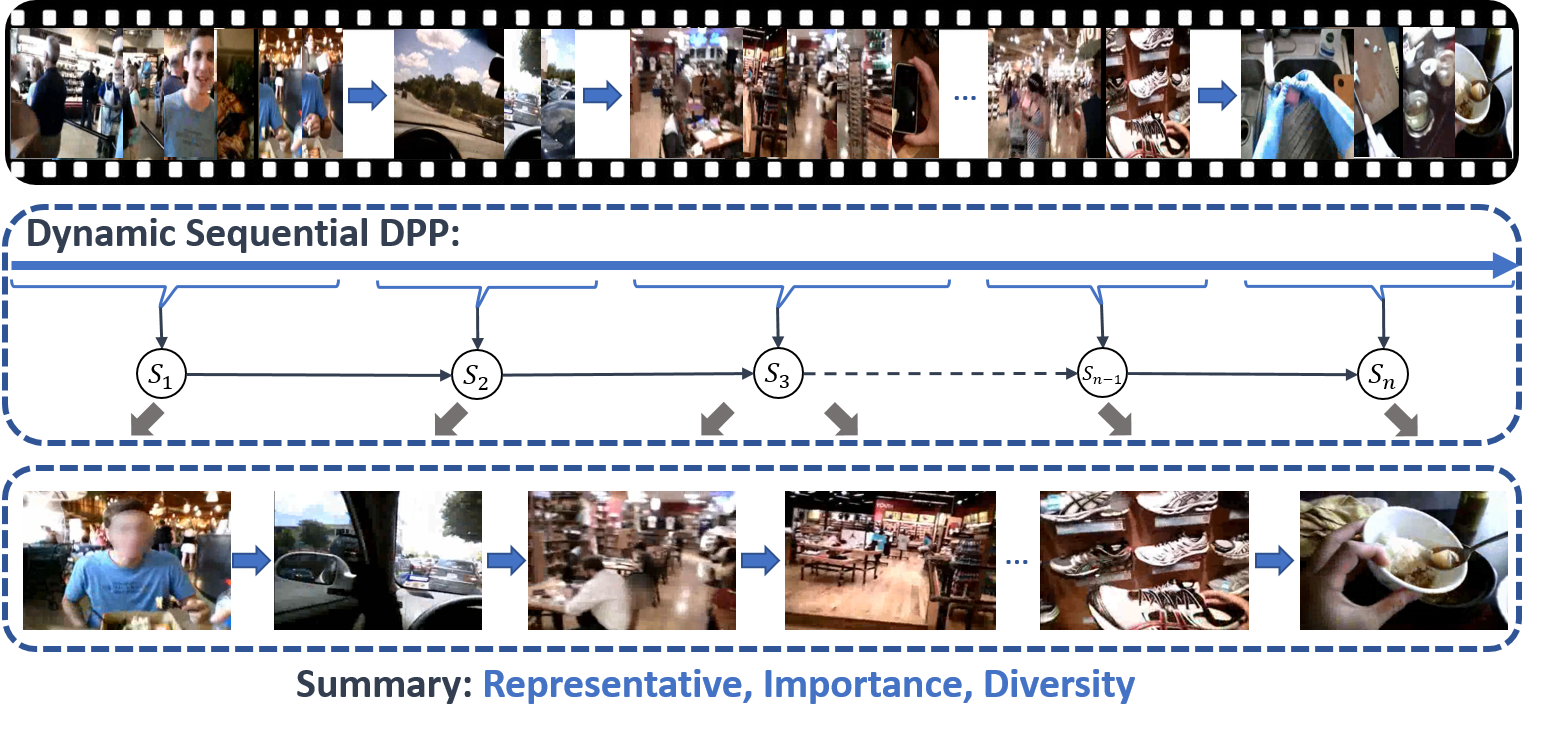}
\vspace{-25pt}
    \caption{Dynamic Sequential DPP (DySeqDPP) for Video Summarization}
    \label{fig:dseqdpp}
    \vspace{-20pt}
\end{figure*}

In the supervised video summarization models, a key factor they are supposed to encompass is the diversity of the selected subset of video shots. This is often imposed by submodularity~\cite{gygli2015video,xu2015gaze} and determinant~\cite{DBLP:conf/nips/GongCGS14,zhang2016summary,sharghi2016query}. When a video sequence is short, \textbf{global diversity} over the whole sequence seems like a natural choice~\cite{zhang2016summary,gygli2015video}.

However, if the videos are lengthy like the egocentric videos that are often hours long, it is necessary to track the temporal structures of the videos and enforce \textbf{local diversity} instead~\cite{DBLP:conf/nips/GongCGS14,sharghi2017query}. The local diversity refers to that the shots selected from a short time duration are diverse but visually similar shots are allowed to co-exist in the summary if they appear far apart in the video. Consider a video sequence that is about ``leaving home for shopping in the morning and then coming back home to have lunch''. Although the video shots of the ``home'' scene in the morning may be similar to those at noon, the summary should contain some shots of both in order to make the summary a complete story carried by the video. 

In this paper, we are mainly interested in summarizing extremely lengthy (e.g., egocentric) videos and, accordingly, models that are capable of observing the \textbf{local diversity}. Among the existing works, sequential determinantal point process (SeqDPP)~\cite{DBLP:conf/nips/GongCGS14} and dppLSTM~\cite{zhang2016video} both account for the temporal dynamics of the videos. However, neither of them explores \emph{``how local''} the {local diversity} should be. Take the SeqDPP for instance, it requires users to manually partition the video into disjoint segments of the same length and then impose diversity both within each of them and between adjacent segments, locally. There is no guiding principle about how to best partition a video sequence into such segments. Besides, it could be sub-optimal to make the segments of the same length because different types of events often unroll at distinct frame rates. The same snags exist in dppLSTM.

We propose to improve the SeqDPP model~\cite{DBLP:conf/nips/GongCGS14} by a latent variable that dynamically controls the time span of a segment upon which the local diversity is then defined in the form of a conditional DPP. In other words, we enable SeqDPP to learn to automatically infer \emph{how local the local diversity is} in the input video. Figure~\ref{fig:dseqdpp} illustrates our main idea. Given an input video shown on the top panel, our dynamic SeqDPP seeks the appropriate and possibly different lengths of the segments (cf.\ the middle panel) from which it selects video shots (the bottom panel) and places them on a story board or links them into a short video clip as the summary of the video. 

Another contribution of this paper is a novel reinforcement learning algorithm for the proposed dynamic SeqDPP (DySeqDPP). While DySeqDPP seems like a straightforward extension to the vanilla SeqDPP, it is less obvious how to efficiently train the model. The DPPs~\cite{DBLP:conf/uai/KuleszaT11} and its variants (e.g., SeqDPP~\cite{DBLP:conf/nips/GongCGS14}, dppLSTM~\cite{zhang2016video}, and SH-DPP~\cite{sharghi2016query}) are almost all trained by the hallmark maximum likelihood estimation (MLE) except for the large-margin DPP~\cite{chao2015large} and Bayesian DPP~\cite{affandi2014learning}. However, it is often difficult to maximize the likelihood of a sequential model with latent variables; gradient ascent fails to track the statistical structure, and the EM algorithm~\cite{dempster1977maximum} becomes involved and inefficient unless one assumes special compositions of a sequential model~\cite{welch2003hidden}. 

In light of these challenges,  we instead provide a reinforcement learning perspective for understanding SeqDPPs. The proposed DySeqDPP is used as a policy by an agent to interact with the environment --- the input video. Accordingly, we train this DySeqDPP model by policy gradient descent~\cite{sutton1998reinforcement}. Not only we do not have to explicitly deal with the latent variables, but also we benefit from the flexible reward functions in policy gradient descent --- we can bridge the training and validation phases of the summarizer by defining the reward function as some evaluation metric(s).

We evaluate this dynamic SeqDPP model on standard video summarization datasets. Extensive results show that it  significantly outperforms competing baselines especially the vanilla SeqDPP, verifying the necessity of dynamically determining how local the local diversity is. The rest of the paper is organized as follows. Section~\ref{sRelated} discusses some related existing video summarization works. 
After that, we describe our dynamic SeqDPP and the reinforcement learning algorithm in Section~\ref{sApproach}. We report empirical results in Section~\ref{sExp} and then conclude the paper by Section~\ref{sConclusion}.
\vspace{-0.1in}
\section{Related Work} \label{sRelated}
\subsection{Video summarization}

Different algorithms for automatic video summarization are generally designed by the same principles. Those informative guidelines contain three main factors: (1) individual interestingness or relevance~\cite{lee2012discovering,lu2013story}, which means selecting  frames/shots that are important in the video; (2) representativeness~\cite{hong2009event,khosla2013large,ngo2003automatic}, which means the summary should contain the main event of the videos; (3) collective diversity or coverage~\cite{liu2002optimization,zhang1997integrated}, which is to reduce redundant frames/shots  without losing much information. These factors are used in most of the existing works. 
Next, we review the representative approaches in two common classes, unsupervised and supervised video summarization.

\textit{Unsupervised video summarization:}
A variety of prior works is designed based on basic visual quality like low-level appearance
and motion cues~\cite{zhang1997integrated,lee2012discovering,ngo2003automatic,lu2013story,hong2009event,khosla2013large,kang2006space,ma2002user,kwon2015unified,liu2002optimization}. Graph models are utilized for event detection in some approaches~\cite{kwon2015unified,ngo2003automatic}.
In general, the criteria applied in those methods for making decisions about including or excluding shots are devised by the system developers empirically. Besides, some approaches leverage Web images for video summarization based on the assumption that the static Web pictures  tend to contain information of interest to people, so the Web images reveal user-oriented importance selecting video shots/frames~\cite{khosla2013large,kim2014joint,xiong2014detecting,chu2015video}.

\textit{Supervised video summarization:}
Recently, several explorations on supervised video summarization have been exerted for various goals~\cite{DBLP:conf/nips/GongCGS14,gygli2015video,zhang2016summary,gygli2014creating,DBLP:conf/uai/ChaoGGS15,lu2013story,lee2012discovering,liu2010hierarchical,sharghi2016query,sharghi2017query,zhang2016video}. They achieve superior performance over the traditional unsupervised clustering algorithms.  Among them, Gygli \textit{et al.} try to add some supervised flavor to optimize mixture objectives with learning each criterion's weight~\cite{gygli2014creating,gygli2015video}.
A hierarchical model has been proposed to learn with few labels, and it is optimized to generate video summary containing interesting objects~\cite{liu2010hierarchical}. Egocentric videos~\cite{del2017summarization} can be compacted with importance of people and objects~\cite{lee2012discovering}; on the other hand, Zheng \textit{et al.} explicitly consider how one sub-event leads to another in order to provide a better sense of story for those kinds of videos~\cite{lu2013story}.
Meanwhile, Yao \textit{et al.} propose a pairwise deep ranking model to highlight video segments of first-person videos~\cite{yao2016highlight}. In conclusion, supervised methods are capable of utilizing the intentions of users about what a qualified video summary is rather than designing the systems only relying on the experts' own perspective.

Besides, as a powerful diverse subset selection model, the determinantal point process (DPP) has been widely used for video summarization.
For instance, Gong \textit{et al.} propose the first supervised video summarization method~\cite{DBLP:conf/nips/GongCGS14} (SeqDPP) as far as we know, it models local diversity to capture the temporal information of videos rather than modeling global diversity.  Combining long short-term memory (LSTM) with DPPs has been studied in~\cite{zhang2016video} to model the variable-range temporal dependency and diversity among video frames at the same time.  Effort has been spent to study transferring summary structures from annotated videos to
unseen test videos in~\cite{zhang2016summary}. Sharghi \textit{et al.} explore the query-focused video summarization in \cite{sharghi2016query,sharghi2017query}. Large margin separation principle has been leveraged for DPPs to estimate parameters in~\cite{DBLP:conf/uai/ChaoGGS15}. 

We will provide more details of DPPs and SeqDPP  in Sections~\ref{dpps} and \ref{sSeqDPP}.
\vspace{-15pt}

\subsection{Reinforcement learning}
\vspace{-5pt}
Sequential models are often optimized by maximizing the likelihood (MLE) of the next ground-truth entry given the previous ground-truth entry during training. However, the model uses the entry from the model's own distribution to predict the next entry during testing. This discrepancy, called \emph{exposure bias}~\cite{ranzato2015sequence}, leads to the error accumulation problem as the sequential model unrolls over time. Another discrepancy is between the likelihood objective used during training and the actual evaluation metrics during testing (e.g., BLEU~\cite{papineni2002bleu} and the F1 score in this work). Since these metric 
are often non-differentiable, it is not immediately clear how to include them in the training objective. 

Reinforcement learning (RL) provides a unified solution to both problems above.
The REINFORCE algorithm~\cite{williams1992simple} is utilized to train recurrent neural network~\cite{ranzato2015sequence}.  Rennie \textit{et al.} borrow ideas from~\cite{ranzato2015sequence} 
in the image captioning task and obtain very promising results~\cite{rennie2016self}. We note that the use of RL in those contexts is icing on the case in the sense that, while RL boosts the results to some degree, the MLE is still applicable. For our DySeqDPP model, however, RL becomes a necessary choice because it is highly involved to handle the latent variables in DySeqDPP by MLE.

\vspace{-0.1in}
\section{Background: DPP and SeqDPP} \label{sBackground}
We briefly review the determinantal point process (DPP) and the sequential DPP (SeqDPP) in this section. It will become clear soon how the former promotes diversity in the selected subsets and the latter enables local diversity.

\vspace{-0.1in}
\subsection{DPPs}\label{dpps}
A discrete DPP defines a distribution over the subsets of a ground set and assigns high probability to a subset if its items are diverse from each other. The notion  of diversity is induced by a kernel matrix whose entries can be understood as pairwise similarities between the items. The more similar two items are, the less likely they co-occur in a subset sampled from the DPP. 

More concretely, given a ground set $\calY=\{1, 2,\ldots,\cN\}$ of $\cN$ items, let $\mK\in\R^{\cN\times \cN}$ be a symmetric positive semidefinite matrix, called the kernel of DPP. It measures pairwise similarities between the $\cN$ items. A distribution over a random subset $Y\subseteq\calY$ is a DPP, if for every $\vy\subseteq\calY$ we have
\begin{align}\label{ePy}
P_{dpp}(\vy\subseteq Y; \mK) = \det(\mK_{\vy}),
\end{align}
where $P_{dpp}(\cdot)$ is the probability of an event, $\mK_{\vy}$ denotes a squared submatrix of $\mK$ with rows and columns indexed by $\vy$, and $\det(\cdot)$ is the determinant of a matrix. All the eigenvalues of the kernel matrix $\mK$ are between 0 and 1. Since $P(i, j\in Y;\mK) = K_{ii}K_{jj}- K_{ij}^2$, {\it i.e.}, the probability of any two items $i,j$ co-existing in the random subset $Y$ is discounted by their similarity $K_{ij}$. In other words, the subsets whose items are less similar to each other are assigned higher probabilities than the other subsets.
\vspace{-0.2in}

\subsubsection{L-ensemble.}
In practice, it is often more convenient to use the so-called L-ensemble DPP that directly assigns atomic probabilities to all the possible subsets of the ground set. Let $\mL$ denote a symmetric positive semidefinite matrix in $\R^{\cN \times\cN}$. The L-ensemble DPP draws a subset $\vy \subseteq \calY$ with probability
\begin{equation}
P_L(Y= \vy ; \mL) = {\det(\mL_{\vy})}/{\det(\mL+\mI)},
\label{eDPPLensemble}
\end{equation}
where $\mI$ is an identity matrix. The corresponding marginal kernel that defines the marginal probability in~(\ref{ePy}) is given by $\mK =  \mL(\mL + \mI)^{-1}$. 


\vspace{-0.1in}
\subsubsection{Conditional DPP.} One of the appealing properties of DPP is that there exists an analytic form of its conditional distribution. For any $\vy_1\subseteq\mathcal{Y}$ and $\vy_0\subseteq\mathcal{Y}$, $\vy_1\cap\vy_0=\emptyset$,
\begin{align}
P_L(Y=\vy_1\cup\vy_0|\vy_0\subseteq Y;\mL) = {\det(\mL_{\vy_1\cup\vy_0})}/{\det(\mL + \mI_{\mathcal{Y}\setminus \vy_0})}, 
\end{align}
where $\mI_{\mathcal{Y}\setminus \vy_0}$ is a matrix with ones in the diagonal entries indexed by $\mathcal{Y}\setminus \vy_0$ and zeros everywhere else.  Kulesza and Taskar have written an excellent tutorial about DPPs~\cite{DBLP:journals/corr/abs-1207-6083}.

\vspace{-0.1in}
\subsection{Sequential DPPs} \label{sSeqDPP}
A sequential DPP (SeqDPP)~\cite{DBLP:conf/nips/GongCGS14} was proposed for supervised video summarization. It adheres to the inherent temporal structure in video sequences, thus overcoming the deficiency of DPPs which treat video frames/shots as randomly permutable items. The main technique is to use the conditional DPPs to construct a Markov chain.

Given a long video sequence $\sV$, we partition it into $\cT$ disjoint yet consecutive short segments $\bigcup_{t=1}^\cT {\cal V}_t = \cal{V}$. At the $t$-th time step, SeqDPP selects a diverse subset of items (e.g., frames or shots), by a variable $X_t\subseteq{\cal V}_t$, from the corresponding segment conditioning on the items $\vx_{t-1}\subseteq{\cal V}_{t-1}$ selected from the immediate past segment. This subset selection variable $X_t$ follows a distribution given by the conditional DPP,
\begin{align}
    P_{seq}(X_t=\vx_{t} | X_{t-1}=\vx_{t-1}, \mathcal{V}_t) \coloneqq& \, P_L(Y_t=\vx_{t}\cup\vx_{t-1} | \vx_{t-1}\subseteq{Y_t}; \mL^t) \\
    = &\, {\det(\mL^t_{\vx_t\cup\vx_{t-1}})} \,/\, {\det(\mL^t + \mI^t_{\mathcal{V}_t})},
    \label{eConditionalDPP}
\end{align}
where $P_L(Y_t;\mL^t)$ is an L-ensemble  with the ground set $\vx_{t-1}\cup \sV_t$. Denote by $\vx_0=\emptyset$. The SeqDPP over all the subset selection variables is factorized as
\begin{equation}
P_{seq}( \{X_t= \vx_t\}_{t=1}^\cT, \mathcal{V}) = \prod_{t=1}^\cT P_{seq}(X_t = \vx_t | X_{t-1}=\vx_{t-1}, \mathcal{V}_t).
\label{eSeqDPP}
\end{equation}

\begin{figure}[t]
\centering
\small
\begin{tabular}{ccc}

\begin{tikzpicture}[
    node distance=1.5em,
    thick,
    terminal/.style={
      draw,circle,inner sep=1pt,minimum size=1.5em
    },
    doubletrans/.style={->,bend left=10}
  ]

\node (s0) [terminal] at (0,0)		{$Y$};
\node (s6) [terminal,above=of s0, fill=lightgray]	{$\calY$};

\foreach \x in {6} {
  \pgfmathparse{int(\x-6)}
  \edef\X{\pgfmathresult}
  \draw [->] (s\x) to node[above] {$$} (s\X);
}
\end{tikzpicture}

~~~~~~~~~~~~

&
\begin{tikzpicture}[
    node distance=1.5em,
    thick,
    terminal/.style={
      draw,circle,inner sep=1pt,minimum size=1.5em
    },
    doubletrans/.style={->,bend left=10}
  ]

\node (s0) [terminal] at (0,0)		{$Y_1$};
\node (s1) [terminal,right=of s0] 	{$Y_2$};
\node (s2) [right=of s1]			{$\cdots$};
\node (s7) [terminal,above=of s1, fill=lightgray] 	{$\calY$};

\foreach \x in {0,1} {
\pgfmathparse{int(\x+1)}
\edef\X{\pgfmathresult}
\draw [->] (s\x) to node[above] {$$} (s\X);
}

\draw [->] (s7) to node[above] {$$} (s0);
\draw [->] (s7) to node[above] {$$} (s1);
\draw [->] (s7) to node[above] {$$} (s2);
\end{tikzpicture}
~~~~~~~~~~~~
&

\begin{tikzpicture}[
    node distance=1.5em,
    thick,
    terminal/.style={
      draw,circle,inner sep=1pt,minimum size=1.5em
    },
    doubletrans/.style={->,bend left=10}
  ]

\node (s0) [terminal] at (0,0)		{$X_1$};
\node (s1) [terminal,right=of s0] 	{$X_2$};
\node (s2) [right=of s1]			{$\cdots$};
\node (s6) [terminal,above=of s0, fill=lightgray]	{$\mathcal{V}_1$};
\node (s7) [terminal,above=of s1, fill=lightgray] 	{$\mathcal{V}_2$};
\node (s8) [above=of s2, fill=lightgray]   {$\cdots$};			

\foreach \x in {0,1} {
\pgfmathparse{int(\x+1)}
\edef\X{\pgfmathresult}
\draw [->] (s\x) to node[above] {$$} (s\X);
}

\foreach \x in {6,7,8} {
  \pgfmathparse{int(\x-6)}
  \edef\X{\pgfmathresult}
  \draw [->] (s\x) to node[above] {$$} (s\X);
}
\end{tikzpicture}

\end{tabular}
\vspace*{-0.1in}
\caption{From left to right: Determinantal point process (DPP)~\cite{DBLP:journals/corr/abs-1207-6083}, Markov DPP~\cite{affandi2012markov}, and sequential DPP (SeqDPP)~\cite{DBLP:conf/nips/GongCGS14}. The ground sets are denoted by the shaded nodes.} \label{fTemporalDPPs}
\vspace{-10pt}
\end{figure}
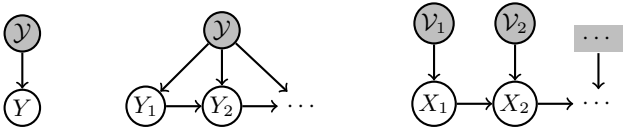

Figure~\ref{fTemporalDPPs} illustrates SeqDPP and compares it to the vanilla DPP and Markov DPP~\cite{affandi2012markov}. Unlike the vanilla or Markov DPPs which considers the video frames/shots as orderless items, SeqDPP maintains the temporal order among the segments and yet ignores it among the frames/shots within an individual segment, locally. Furthermore, it retains the diversity property for adjacent video segments but not for those that are far apart. Indeed, users may want to keep visually similar video clips in the summary if they are far apart in a lengthy video in order to tell a complete story of the video. 

\vspace{-0.1in}
\subsection{Reinforcement learning}
Consider an agent that takes actions according to some policy to interact with the environment. Following the popular Markov decision process (MDP) formalism, we describe the problem  by $(\mathcal{S},\mathcal{A},P,R,\gamma)$, where $\mathcal{S}$ and $\mathcal{A}$ are the state $(s)$ space and action $(a)$ space, respectively, $P(s_{t+1}|s_t,a_t)$ is a state transition distribution, $R(s_{t+1};s_t,a_t)$ is a reward the agent receives if it takes action $a_t$ at state $s_t$ and results in state $s_{t+1}$, and $\gamma\in(0,1)$ is a discount factor. A policy is denoted by $\pi:\mathcal{S}\mapsto\mathcal{A}$, which is essentially a conditional distribution $\pi(a_t|s_t)$ over the actions given any state. Reinforcement learning aims to find the agent a policy that maximizes the expected total discounted reward $\mathbb{E}_\pi \sum_{i=0}^\infty \gamma^i R_{t+i}$ starting from time step $t$. 
\vspace{-0.1in}
\section{Reinforcing dynamic SeqDPPs} \label{sApproach}
We are now ready to present our dynamic SeqDPP (DySeqDPP) along with a reinforcement learning algorithm for estimating the model parameters.

\vspace{-0.1in}
\subsection{DySeqDPP}
We describe the DySeqDPP model using the MDP formalism $(\mathcal{S},\mathcal{A},P,R,\gamma)$ so that the corresponding learning algorithm follows naturally. We note that, in addition to the new DySeqDPP, another contribution of this section is the reinforcement learning perspective for understanding SeqDPPs. Under this framework, SeqDPP and DySeqDPP can be seen as two types of stochastic policies. 
\vspace{-0.05in}

\begin{description}
\item[State $s_t$ at time step $t$:] An information state  is about the history of an agent's observations (and rewards) about the environment. It is used to determine what happens next upon an action taken by the agent.
In our context, the state $s_t=\{\bigcup_{t'=1}^{t-1}\vx_{t'}, \mathcal{V}_t\}$ comprises the dynamic partition of the video  $\mathcal{V}_{t}$ at time step $t$ and the generated video summary $\bigcup_{t'=1}^{t-1}\vx_{t'}$ right before the current step $t$. One may wonder to alternatively treat all the video segments $\mathcal{V}_{1},\cdots,\mathcal{V}_{t}$ until step $t$ as the state. We contend that it is oppressive and unnecessary to carry them along over time. Instead, the summary of the past conveys similar amount of information by design.


\item[Action $a_t$ at time step $t$:] In DySeqDPP, the agent takes actions 1) to select a subset $X_t$ from the video segment $\mathcal{V}_t$ and 2) to propose the length $L_t$ of the next segment $\mathcal{V}_{t+1}$. The subset selection variable $X_t\subseteq\mathcal{V}_t$ and the partition proposal variable $L_t\in\mathcal{L}$ jointly define the action space. In other words, an action takes the form of $A_t=(X_t,L_t)$ whose realization is denote by $a_t=(\vx_t,l_t)$. We limit the search of the segment's length to the range of $\mathcal{L}=\{5,6,\cdots,15\}$  shots.


\item[Policy $\pi$:] We let the agent take a stochastic policy in the following manner, \begin{align}
\pi(a_t|s_t) = P(\vx_t,l_t|s_t)= &P(\vx_t|s_t)P(l_t|\vx_{t},s_{t}),\label{ePolicy}
\end{align}
where $P(\vx_t|s_t)$ is a conditional DPP used to build SeqDPP~\cite{DBLP:conf/nips/GongCGS14}, {\it i.e.},
\begin{align}
P(\vx_t|s_t) =&\, P(\vx_t|\cup_{t'=1}^{t-1}\vx_{t'}, \mathcal{V}_{t})
\coloneqq\, P_L(Y_t=\vx_t\cup\vx_{t-1}|\vx_{t-1}\subseteq Y_t; \mL^t) \label{ePxs}
\end{align}
and $P(l_t|\vx_t,s_t)$ is defined as a softmax function,
\begin{align}
P(l_t|\vx_t,s_{t}) = P(l_t|\cup_{t'=1}^t\vx_{t'},\mathcal{V}_{t}) \coloneqq \texttt{softmax}(\vw_{l_t}^T\phi(\cup_{t'=1}^{t}\vx_{t'},\mathcal{V}_{t})). \label{ePlxs}
\end{align}

There are several points in the above worth clarifying and discussing. First of all, equations~(\ref{ePolicy}--\ref{ePlxs}) describe the main body of our DySeqDPP model. It improves SeqDPP by the partition proposal variable $L_t$. It is a latent variable because users annotate summaries of videos without explicitly knowing the boundaries of the local diversities they have in their minds. Secondly, we condition the DPP in eq.~(\ref{ePxs}) on its immediate past time step ($\vx_{t-1}$) only instead of the whole history of summaries included in the state $s_t$. This is due to the same modeling intuition as SeqDPP, {\it i.e.}, in order to maintain local diversity in the summaries. Thirdly, $\phi(\cdot)$ in eq.~(\ref{ePlxs}) extracts features by max-pooling the  representations of all the video shots in the current state $s_t$ as well as the new summary $\vx_t$ selected according to eq.~(\ref{ePxs}). This ensures that sufficient information about both the whole past history and the current of the video is supplied to the {\tt softmax} for the agent to predict the appropriate length of the next segment. Last but not the least,  $\{\vw_{l},l\in\mathcal{L}\}$ are the model parameters to be learned from the user-annotated summaries. It is important to note that the parameters are not bound to any particular environments/videos at all, so the policy can be generalized to unseen videos, too. We postpone the parameterization of the L-ensemble DPP's kernel $L$ to Section~\ref{sLearning}.

\item[State-action value function:] Our goal is to learn a policy to maximize the expected total discounted reward the agent receives, called the state-action value function,
\begin{align}
    &Q^{\pi}(s_0,a_0)\coloneqq\mathbb{E}_{\pi}\Big[\sum_{t=0}^{T}g(\gamma,t)R_t|S_0=s_0,A_0=a_0\Big], 
    \label{eValueFunc}
\end{align}
where $g(\gamma,t)\in[0,1]$ is a discount function and the reward $R_t=R(s_{t+1};s_t,a_t)$ is a function of the state and action. For video summarization, the reward can be evaluation metrics like precision, recall, or F-score computed between the video shots $\cup_{t'=1}^t\vx_{t'}$ selected by the agent and the user summaries of the video (until the current segment $\mathcal{V}_t$). The total number of time steps the agent can take is $T$, which satisfies $\sum_{t=0}^{T-1}l_t<|\mathcal{V}|$ and $\sum_{t=0}^{T}l_t\ge|\mathcal{V}|$.
\vspace{-0.05in}
\end{description}

It is import to note that our goal is to maximize the state-action value function at the initial state and action $(s_0,a_0)$ which are fixed to $s_0=\emptyset$ and $a_0=(\vx_0=\emptyset,l_0=10)$ in our experiments. In  contrast to conventional setups in reinforcement learning, we do not care about the  state-action values at other states because only the initial state gives rise to a whole summary of the video, which is our interest. This insight also suggests a special design of the discount function $g(\gamma,t)$. Instead of using the common practice $\gamma^t$, we let it be $g(\gamma,t)=\gamma^{|\mathcal{V}-t|}, \gamma\in(0,1),$ monotonically increasing with respect to $t$ in order to weigh the reward of the whole summary more than the incomplete summaries at any other time steps. 

Those differences highlight the fact that video summarization actually lacks some  characteristics of reinforcement learning (e.g., delayed feedback). Hence, we have to customize the MDP formalism in order to match it with the goal of interest. Nonetheless, by casting DySeqDPP as a policy, we can conveniently learn its model parameters by algorithms in reinforcement learning --- we employ gradient descent in this paper. 

\vspace{-0.1in}
\subsection{Policy gradient descent for learning DySeqDPP} \label{sLearning}
We review the model parameters in DySeqDPP before deriving the learning algorithm. We parameterize two conditional distributions in DySeqDPP for the purpose of out-of-sample extension, so that one can readily apply the learned model to  unseen test videos. The first is the partition proposal distribution (eq.~(\ref{ePlxs})) and the second is the conditional DPP (eq.~(\ref{ePxs})) at each time step $t$, whose L-ensemble kernel is constructed as follows,
\begin{align}
[\mL^t]_{ij} = \vz_i^T\mW^T\mW\vz_j, \qquad \vz_i =  \texttt{ReLU}(\mU\, \texttt{ReLU}(\mV \vf_i)) \label{ePara}
\end{align}
where $\vf_i$ is the feature representation of video shot $i$ in the ground set $\vx_{t-1}\cup\mathcal{V}_t$ of the time step $t$. This feature vector goes through a feedforward network with \texttt{ReLU} activations. Denote by $\theta$ the union of the weights of the network $(\mW,\mU,\mV)$ and the unknowns $\{\vw_l,l\in\mathcal{L}\}$ in eq.~(\ref{ePxs}). We next derive a learning algorithm using the policy gradient descent~\cite{sutton-1998-reinforcement} to estimate the model parameters $\theta$.

Recall that our goal is to maximize the state-action value function at the initial state and action. Denoting by $J\triangleq -Q^\pi(s_0,a_0)$, we can minimize it by gradient descent,
\begin{align}
\nabla_\theta J |_{\theta=\theta_{\text{old}}} &= - \mathbb{E}_{\bm{\tau}\sim\pi(\theta_{\text{old}})}\Big[\sum_{t=1}^{T}g(\gamma,t)R_t\nabla_\theta\log p(\bm{\tau};\theta)|_{\theta=\theta_{\text{old}}}\Big] \\
&\approx - \frac{1}{K}\sum_{k=1}^K\Big[\sum_{t=1}^{T_k}g(\gamma,t)\,r_{tk}\,\nabla_\theta\log p(\bm{\tau}_k;\theta)|_{\theta=\theta_{\text{old}}}\Big] \label{eApprox}
\end{align}
where the last equation is obtained by sampling $K$ trajectories $\{\bm{\tau}_k\}$ from the policy instantiated by the old parameter $\theta_\text{old}$, $r_{tk}$ is the reward that the agent receives at time step $t$ of the $k$-th trajectory, and the first equation is due to the following fact,
\begin{align}
\nabla_\theta \mathbb{E}_{x\sim\theta}[f(x)] |_{\theta=\theta_{\text{old}}} &= \mathbb{E}_{x\sim\theta_{\text{old}}}\Big[\nabla_\theta\log p(x;\theta)|_{\theta=\theta_{\text{old}}}f(x)\Big].
\end{align}

We still need to work out $\nabla_\theta\log p(\bm{\tau};\theta)$ in eq.~(\ref{eApprox}). The key is that the state-transition distribution $p(s_{t+1}|s_{t},a_{t})$ is actually deterministic under our context laid out in Sec.~\ref{sApproach} (because the action $a_t$ fully determines the summary $\vx_{t}$ and the next segment $\mathcal{V}_{t+1}$, and hence the next state). Therefore, for a  trajectory $s_0,a_0,s_1,a_1,\cdots$, we have

\vspace{-0.2in}
\begin{align}
& \nabla_\theta\log p(\bm{\tau};\theta) 
 =  \nabla_\theta\log\Big[ p(s_0,a_0)\prod_{t=1}^{T} p(s_t|s_{t-1},a_{t-1})\pi(a_t|s_t;\theta)\Big] \\
= \,& \nabla_\theta \sum_{t=1}^{T} \log\pi(a_t|s_t;\theta) =\, \sum_{t=1}^{T} \Big[\nabla_{\theta}\log P(\vx_t|s_t) +  \nabla_{\theta} \log P(l_t|\vx_t,s_t)\Big]
\end{align}
where the first summand of the last equation is the gradient with respect to the parameters of conditional DPP and the second is of the \texttt{softmax} (eq.~(\ref{ePlxs})). 


\eat{
\begin{align}
\nonumber \nabla_\theta J \sim - \sum_{t'=1}^{|\mathcal{V}|} r_{t'} \sum_{t=1}^{|\mathcal{V}|} \Big[\nabla_{\mW}\log P(\vx_t|\vx_{t-1},\mathcal{V}_t) \\ +  \nabla_{\vw_l} \log \text{softmax}(\vw_l^T\phi(\bigcup_{t'=1}^{t-1}\vx_{t'},\mathcal{V}_{t}))\Big]
\end{align}
}

\vspace{-0.15in}
\paragraph{Implementation:} Instead of computing the gradients explicitly, one may use the ``autodiff'' feature of many existing deep learning tools to obtain the gradients. Take \textsc{pytorch} (\url{http://pytorch.org}) for instance. We may program the following for a trajectory, 
\begin{align}
\nonumber J(\bm{\tau};\theta) = - \sum_{t=1}^{T} g(\gamma,t)\,r_t\,\Big[\log P(\vx_t|s_t;\theta)  + \log P(l_t|\vx_t,s_t;\theta)\Big],
\end{align}
and then use the \texttt{backward()} function to automatically compute the gradients followed by calling the \texttt{step()} function to do a one-step gradient descent. After that, we sample another trajectory and repeat the procedure until  the termination condition.

\vspace{-0.1in}
\section{Experiments}\label{sExp}
\vspace{-0.1in}
We run experiments on three datasets, UTE~\cite{lee2012discovering}, SumMe~\cite{gygli2014creating}, and TVSum~\cite{song2015tvsum}, and compare our approach to several competing baselines. 

\vspace{-0.1in}
\subsection{The UT egocentric (UTE) dataset}\label{uteEx}
\vspace{-0.05in}
\subsubsection{Data and features.}
UTE~\cite{lee2012discovering} contains four egocentric videos, each of which lasts between three and five hours long. It captures daily activities such as shopping in a grocery store, having lunch, working, chatting with friends, meeting with colleagues, etc. In addition to the big variety of content, the videos are also quite challenging due to ego motions --- as a result, the views change frequently. The motion blur is more frequent and severe than ``third-person'' videos. In general, the video shots of an activity are placed in between of blurred frames and nuisance views. Following the experiment protocol of~\cite{sharghi2017query}, we run four rounds of experiments. In each round, we use two videos for training, one for validation, and the last for testing. We uniformly divide the videos to 5-second shots. From each video frame, we extract 4,096D deep CNN features as the activation of the last fully connected layer of the VGG19 network~\cite{simonyan2014very} pretrained on ImageNet~\cite{imagenet_cvpr09}. After that, we use PCA to reduce the feature dimension from 4,096D to 512D, followed by max-pooling within each shot in order to have a shot-level feature representation ({\it i.e.}, $\vct{f}_i$ in eq.~(\ref{ePara})).

\vspace{-0.15in}
\subsubsection{Competing methods.} 
We mainly compare our approach (DySeqDPP) to the following methods and their variations which, like ours, locally promote diversity in video summaries: SeqDPP~\cite{DBLP:conf/nips/GongCGS14,lu2013story}, dppLSTM~\cite{zhang2016video}, and uniform sampling (Uniform). We let the methods automatically work out the lengths of the summaries except for the uniform sampling, to which we supply the lengths of the oracles. For SeqDPP, however, the length of each segment has to be manually set. In addition to the 10-shot segments suggested in the original work~\cite{DBLP:conf/nips/GongCGS14}, we also include the results of segments of 5 shots and 12 shots. Finally, we include another comparison by improving the original SeqDPP with our reinforcement learning algorithm. This is implemented by fixing the partition proposal distribution $P(L_t|\vx_t,s_t)$ as a Dirac delta function $\delta(L_t=l)$, where $l=10$ is independent of the time steps. Besides, we  learn using the reward of the whole summary by setting $g(\gamma,t)=0$ for $t<T$ and $g(\gamma,T)=1$, unless specified otherwise.

\vspace{-0.15in}
\subsubsection{Evaluation.}
In the literature, system-generated summaries have been evaluated in a variety of manners including but not limited to user studies~\cite{lee2015predicting}, percentage of frames overlapped with user summaries~\cite{zhang2016video}, bipartite matching based on distances of low-level visual features~\cite{sharghi2017query}, etc. Arguably, user study is the ``gold'' standard, but it is extremely time-consuming. In this paper, we instead use the bipartite matching based on a ``semantic distance'' --- pairwise Hamming distance between video shots computed upon the concepts annotated for each shot. This imitates user studies  in the sense that the ``semantic distance'' is strongly correlated with users' perceptions about the difference between a system-generated summary and an actual user's summary. The concepts per video shot are borrowed from an earlier work by Sharghi et al.~\cite{sharghi2017query}, in which the authors asked users to choose from 54 concepts the ones relevant to a given video shot. 

Given two summaries ({\it i.e.}, a system-generated one and a user summary), we construct a bipartite graph between them with the shots as nodes. A node in one part is connected to all the nodes in the other part with edge weights as the (negative) Hamming distance computed from the per-shot concepts~\cite{sharghi2017query}. After that, we find the size of the maximum bipartite matching and divide it by the length of the user (system) summary to obtain the recall (precision). Additionally, we improve this metric by removing the edges between the video shots that are more than $K$ time steps away from each other. In other words, if two shots are far away from each other for more than $5K$ seconds, there is no edge between them in the improved evaluation metric.

\begin{figure}[t]
  \centering
  \includegraphics[width = 0.7\textwidth]{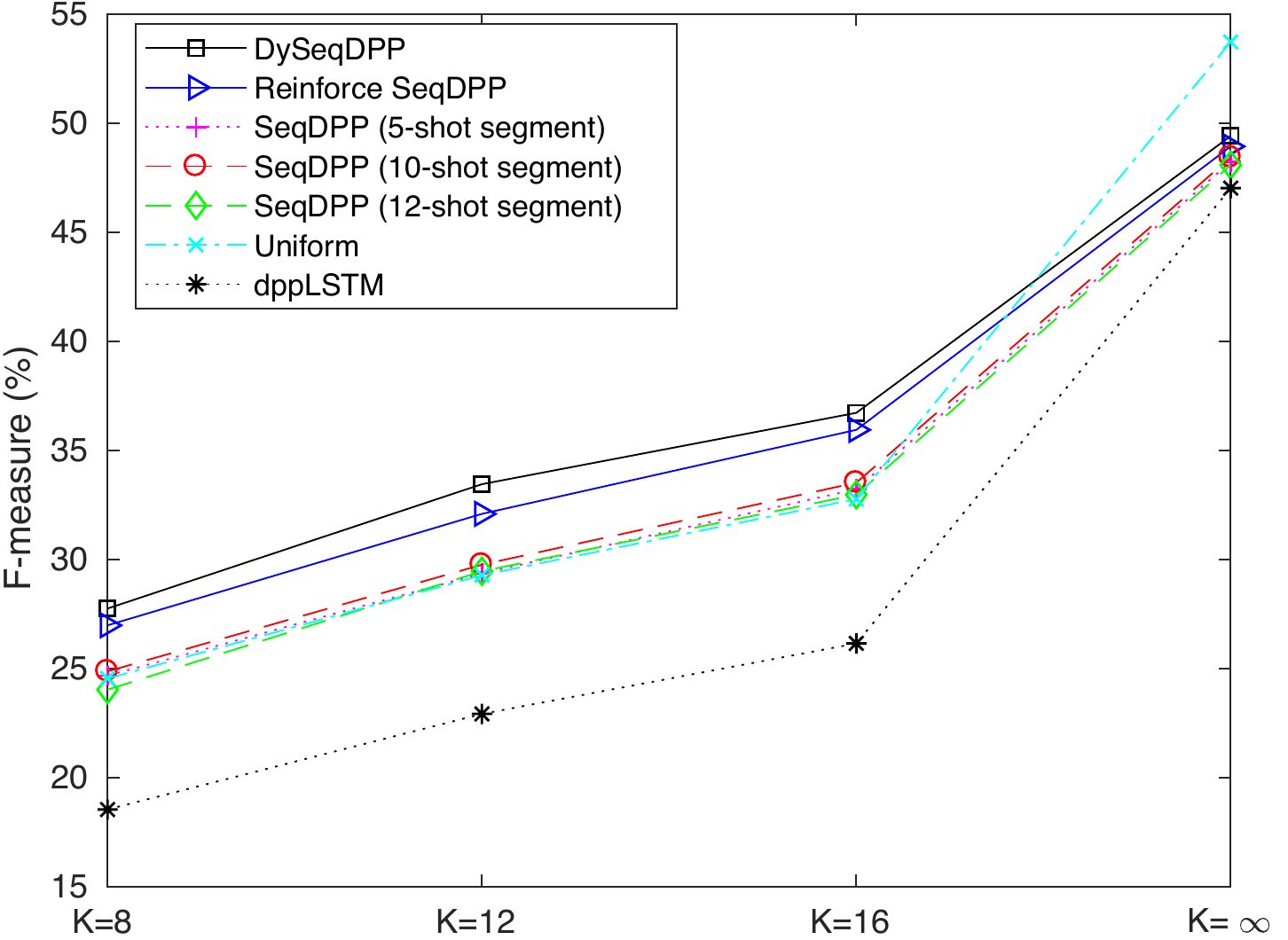}
\vspace{-10pt}
  \caption{Comparison results in terms of average F1-score (the higher, the better) for 4 videos in UTE dataset where horizon axis means different $K$ used in local bipartite matching.}
  \label{fig:ute_perf_final}
  \vspace{-15pt}
\end{figure}

\vspace{-0.20in}
\subsubsection{Comparison results.}

Figure~\ref{fig:ute_perf_final} 
reports the results using the above evaluation scheme at $K=8,12,16, \infty$. Each system-generated summary is compared against three user summaries and the corresponding precision, recall, and F-measure scores are averaged to reduce  user bias. We can see that the proposed DySeqDPP outperforms the competing methods by a large margin. The SeqDPP trained by our reinforcement learning algorithm ranks the second. These results verify the benefit of understanding the SeqDPPs from the novel reinforcement learning perspective. Moreover, the latent variable for dynamically partitioning the videos into segments also helps. It not only removes the need of handcrafting the segments but also gives rise to superior performance over the equally paced segments. 

Another intriguing observation is that there is no significant difference among the results of SeqDPP when we change the sizes of the segments ({\it i.e.,} 5, 10, and 12 shots). It indicates that one can hardly find an ``optimal'' length for the equally placing segments of SeqDPP, signifying the need of dynamically partitioning the videos to segments of variable lengths as our DySeqDPP does. 

It is a little surprising to see that dppLSTM underperforms uniform sampling. Upon examining the existing works~\cite{sharghi2017query,DBLP:conf/eccv/Gong16} carefully, we find that uniform sampling is actually a very competitive baseline partially because it receives unfair information at inference --- length of the oracle summary. Another possible reason is that we did not pre-train the dppLSTM using any additional datasets as done in~\cite{zhang2016video}. 

\vspace{-0.1in}
\begin{table}[t]
\centering
\caption{Comparison results on  UTE  evaluated by the bipartite matching F1-score ($K=12$)} \label{table:resultsute_final}
\vspace{-5pt}
\begin{tabular}{l|l|l|l|l|l|l|l|l|l|l}
\hline
Method           &                   &        Full      & Full        &Full           &      Partial  & Partial      & Partial      & Greedy    & Pool   & Pool   \\
                 & $\gamma=1e^{-20}$ &$\gamma=0.2$      &$\gamma=0.5$ &$\gamma=0.9$   & $\gamma=0.2$  & $\gamma=0.5$ & $\gamma=0.9$ & Sample    & Seg    & Video   \\
\hline
video 1          & 29.53          &28.96        & 28.03       & 29.27           &  28.83              &  28.33             &  28.23   & 27.76                 & 29.19    & 30.33            \\ \hline
video 2          & 31.17          &30.67  & 31.61       & 30.80                 &  32.53              &  32.07             &  30.91   & 29.24                & 31.20    &  31.90     \\ \hline
video 3          & 46.38          &45.79  & 45.88       & 42.04                 &  45.20              &  45.23             &  44.42   & 43.56               & 40.40    &  43.68\\ \hline
video 4          & 26.72          &26.93  & 26.35       & 26.51                 &  26.07              &  26.41             &  27.51   & 23.81                 & 24.56    &  24.91\\ \hline
Avg.             & 33.45          &33.08  & 32.96       & 32.16                 &  33.15              &  33.01             &  32.77   & 31.09               & 31.33    &  32.71\\ \hline

\end{tabular}
\vspace{-10pt}
\end{table}

\eat{
\vspace{-0.1in}
\begin{table}[t]
\centering
\caption{Comparison results on video summarization on UTE dataset with local bipartite matching (K=12). The results are evaluated by F1-score, the higher the better.} \label{table:resultsute_final}
\begin{tabular}{l|l|l|l|l|l|l|l|l|l|l}
\hline
Method           & full              &        full & full        &full         &      partial & partial      & partial     & Greedy  & Pool  & Pool   \\
                 & $\gamma=1e^{-20}$ &$\gamma=0.2$ &$\gamma=0.5$ &$\gamma=0.9$ & $\gamma=0.2$ & $\gamma=0.5$ & $\gamma=0.9$ & Sample & Seg  & Video   \\
\hline
video 1          & 29.53          &28.96        & 28.03       & 29.27       & 27.76         & 29.19    & 30.33            \\ \hline
video 2          & 31.17          &30.67  & 31.61       & 30.80       & 29.24         & 31.20    &  31.90     \\ \hline
video 3          & 46.38          &45.79  & 45.88       & 42.04       & 43.56         & 40.40    &  43.68\\ \hline

video 4          & 26.72          &26.93  & 26.35       & 26.51       & 23.81         & 24.56    &  24.91\\ \hline
Avg.             & 33.45          &33.08  & 32.96       & 32.16       & 31.09         & 31.33    &  32.71\\ \hline

\end{tabular}
\end{table}
\vspace{-0.1in}
}

\vspace{-0.05in}
\subsubsection{Ablation study.} 
Besides, we run some ablation studies to test several variations to our approach and illustrate the quantitative results in Table~\ref{table:resultsute_final}. First, instead of sampling $K$ trajectories $\{\bm{\tau}_k\}$ based on the old policy, we sample the trajectory $\bm{\tau}$ in a greedy manner, which chooses the subsets with the maximum probability at each step during training. The ``Greedy Sample" column in Table \ref{table:resultsute_final} indicates that greedy sampling produces worse video summarization results. The reason is that the system can not explore the real environment (video) thoroughly under the greedy sampling strategy. 

We also study how the hyper-parameter $\gamma$ ($\gamma=1e^{-20},0.2,0.5,0.9$) influences the model. Specifically,
larger $\gamma$ means we give higher weight to the incomplete summaries at early time steps. Meanwhile $\gamma=1e^{-20}$ means we just consider the whole video summary at the final time step. The experimental results in Table \ref{table:resultsute_final} verify our intuitive assumption that weighing more on the reward of the whole summary is better than on the incomplete summaries at other time steps. In addition, we notice a problem that it is unreasonable to calculate the reward of each time step by comparing the incomplete summary up to the current step with the full user summary (shown in the columns titled ``Full $\gamma=0.2/0.5/0.9$"). To address this problem, we compute the reward by comparing the current system summary with the user summary until this time step, as shown in the column titled ``Partial $\gamma=0.2/0.5/0.9$". The experimental results verify that the latter kind of reward calculation is more reasonable.

Finally, we also study what features work better for predicting $l_t$. Recall that, for $\phi(\cup_{t'=1}^{t}\vx_{t'},\mathcal{V}_{t})$, we concatenate the features of the generated video summary until the current time step and the features of the current segment. We test two alternatives. One is pooling the features of this segment only (PoolSeg) and the other is pooling the features of the whole video sequence up to the current segment (PoolVideo). PoolSeg gives rise to worse results than PoolVideo since it lacks the larger context than the current segment only. PoolVideo is a little worse than and certainly incurs more computation cost than  $\phi(\cup_{t'=1}^{t}\vx_{t'},\mathcal{V}_{t})$  because pooling over the video encounters  redundant information.

\vspace{-10pt}
\subsection{The SumMe and TVSum datasets}\label{SumEx}
\vspace{-0.05in}
\subsubsection{Experiment setup.} In addition to the egocentric videos, we also test our approach on two other popular datasets for video summarization: \textbf {SumMe}~\cite{gygli2014creating} and \textbf{TVSum}~\cite{song2015tvsum}. They are both ``third-person'' video datasets. SumMe consists of 25 consumer videos covering holidays, events, and sports. The lengths of the videos range from about one to six minutes. TVSum contains 50 videos of 10 categories downloaded from YouTube. The videos are one to five minutes in length.

\begin{figure*}[t]
  \centering
  \includegraphics[width = 1\textwidth]{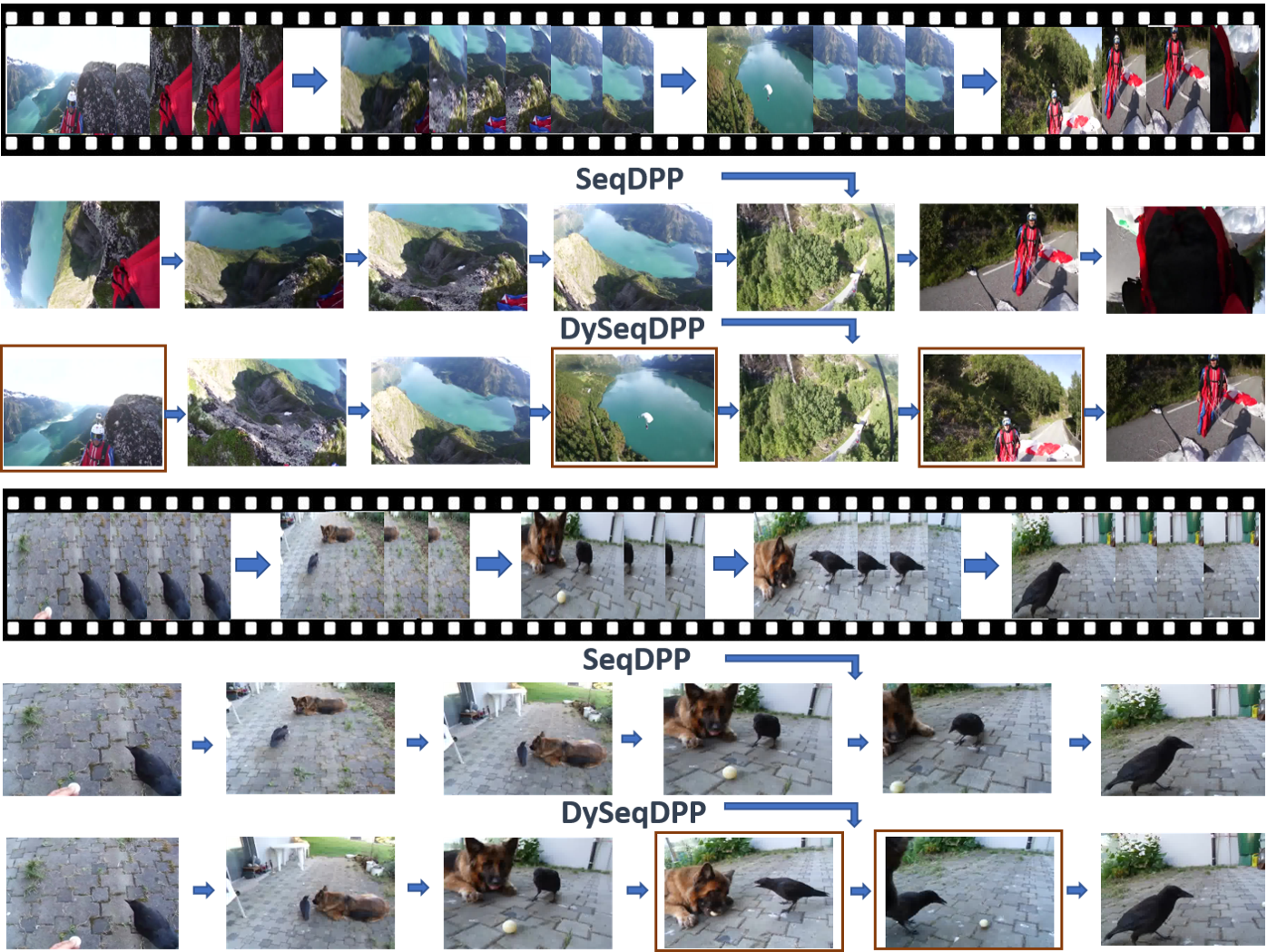}
   \vspace{-10pt}
  \caption{Generated video summary examples with SeqDPP and DySeqDPP}
  \label{fig:showresult}
  \vspace{-10pt}
\end{figure*}

We follow the same experimental setup as dppLSTM~\cite{zhang2016video} in this work. We extract the output (1,024D) of the penultimate layer (pool 5) of  GoogLeNet~\cite{Szegedy_2015_CVPR}  for each video frame. Followed by max-pooling within each shot (15 frames), we get the shot-level feature representation.
In our experiments, we train the model with $60\%$ videos of SumMe (TVSum) , validate on $20\%$ of the dataset, and test on the remaining $20\%$ videos. We run 10 rounds of experiments with different random splits of the dataset and report both the mean F1-scores and standard errors.

\vspace{-0.15in}
\subsubsection{Evaluation.}  

\begin{table}[t]
\centering
\caption{Comparison results on video summarization on SumMe and TVsum dataset. The results are evaluated by F1-score, the higher the better. 
}
\label{table:resulttvsum}
\begin{tabular}{l|l|l|l}
\hline
Dataset                & Method     & Unsupervised & Canonical \\\hline
\multirow{9}{*}{SumMe} & Video-MMR~\cite{li2010multi}          & 26.6         &           \\ \cline{2-4} 
                      & Gygli \textit{et al.}~\cite{gygli2014creating}          &         & 39.4            \\ \cline{2-4} 
                      & Gygli \textit{et al.}~\cite{gygli2015video}           &         & 39.7           \\ \cline{2-4} 
                      & Zhang \textit{et al.}~\cite{zhang2016summary}           &         &  40.9          \\ \cline{2-4} 
                      & vsLSTM~\cite{zhang2016video}    &              & $37.6 \pm 0.8$     \\ \cline{2-4} 
                      & dppLSTM~\cite{zhang2016video}   &              & $38.6 \pm 0.8$     \\ \cline{2-4} 
                      & SeqDPP~\cite{DBLP:conf/nips/GongCGS14}    &              & $40.8 \pm 4.8$     \\ \cline{2-4} 
                      & \textbf{DySeqDPP} &              & $\bm{44.3 \pm 2.8}$     \\ \hline
\hline
\multirow{8}{*}{TVSum}& LiveLight~\cite{zhao2014quasi}           & 46.0         &           \\ \cline{2-4} 
                      & Khosla \textit{et al.}~\cite{khosla2013large}         & 36.0         &           \\ \cline{2-4} 
                      & Song \textit{et al.}~\cite{song2015tvsum}           & 50.0         &           \\ \cline{2-4} 
                      & vsLSTM~\cite{zhang2016video}      &              & $54.2 \pm 0.7$      \\ \cline{2-4} 
                      & dppLSTM~\cite{zhang2016video}    &              & $54.7 \pm 0.7$     \\ \cline{2-4} 
                      & SeqDPP~\cite{DBLP:conf/nips/GongCGS14}   &              & $57.4 \pm 2.0$    \\ \cline{2-4} 
                      & \textbf{DySeqDPP} &              & $\bm{58.4 \pm 2.5}$     \\ \hline
\end{tabular}
\vspace{-5pt}
\end{table}

We evaluate the results again by F1-score. However, the precisions and recalls for computing the F1-score are calculated in a different way from the bipartite graph matching earlier. Following by the practice in dppLSTM~\cite{zhang2016video}, we first split a video into a set of disjoint temporal scenes (which are usually longer and contain more visual information than the segments and shots used in the UTE dataset) using the KTS approach~\cite{potapov2014category}.
We train the model with shot-level feature representations and then use it to obtain shot-level importance scores. Specifically, the importance score of each frame is equal to the score of shots they belong to. We compute the scene-level scores by averaging the scores of frames within each scene and then rank the scenes in the descending order by their scores. In order to generate a video summary, we select the scenes with a duration below a certain threshold ({\it e.g.,} using the knapsack algorithm as in~\cite{song2015tvsum}).
\eat{
They first temporally segment a video into disjoint
intervals using KTS~\cite{potapov2014category} and then they got the shot-level importance score predicted by model. For each frame, their importance score is equal to the score of shot they belong to. Furthermore, they compute interval-level scores by averaging the scores of frames within each interval and then rank intervals in
the descending order by their scores. They select them in order to form a video summary so that the total
duration of video summary is below a certain threshold (e.g., using the knapsack algorithm
as in~\cite{song2015tvsum}).
}Finally, we calculate the precision, recall, and  F1-score according to the temporal overlap between the generated  summary and the user summaries. 

In order to account for the above evaluation scheme, we make some changes to our reinforcement learning algorithm on these two datasets. For training process, firstly we sample the partition proposal $l_t$ with oracle summary based on the old policy on each time step. Thus we can utilize the diagonal values of $\mL^t$ as shot-level scores and then generate the video summary using the approach described above.
Consequently, we can get the reward (F1-score) with the generated video summary. Note that the trajectory $\bm{\tau}$ here is the oracle summary. Therefore, we can optimize the dynamic SeqDPP with reinforcement learning.

\vspace{-0.15in}
\subsubsection{Comparison results.}

Table~\ref{table:resulttvsum} shows the comparison results between our DySeqDPP and several baselines. Note that some of the baseline methods are unsupervised so they are tuned to achieve the best results on the test set. Nonetheless, the supervised ones in general perform better than them. Both SeqDPP and DySeqDPP significantly outperform the others and DySeqDPP ranks to the first by a big margin on SumMe.

\vspace{-0.15in}
\subsubsection{Qualitative results.}

Figure~\ref{fig:showresult} demonstrates some exemplar video summaries generated by SeqDPP and our DySeqDPP, respectively. It is interesting to see that DySeqDPP captures some shots that are key for the story flow and are yet missed by SeqDPP. Take the first video for instance. The sky diver shows up only at the end of SeqDPP's summary while s/he is kept at both the beginning and the end of DySeqDPP's summary. The second is an amusing video recording how a bird saves a ball from a dog's mouth. However, SeqDPP fails to select the key shot in which the dog bites the ball. 

\vspace{-0.1in}
\section{Conclusion} \label{sConclusion}
\vspace{-0.1in}
In this paper, we study \emph{``how local''} the {local diversity} should be for video summarization and utilize it as a guideline to devise a sequential model to tackle the dynamic diverse subset selection problem. Furthermore, we apply reinforcement inference~\cite{sutton1998reinforcement} in the dynamic seqDPP model to solve the problem of \emph{exposure bias} ~\cite{ranzato2015sequence} as well as the issue of non-differentiable metrics existing in SeqDPP~\cite{DBLP:conf/nips/GongCGS14}. The proposed DySeqDPP can not only seek the appropriate and possibly different lengths of segments dynamically, but also bridge the training and validation phases. Experimental results on video summarization demonstrate the effectiveness of our approach. 

\paragraph{Acknowledgements.} {This work was supported in part by NSF IIS 1741431 \& 1566511, gifts from Adobe, and gift GPUs from NVIDIA.} B.G.\ would like to thank Trevor Darrell, Charless Fowlkes,  Alexander Ihler, Dequan Wang, and Huazhe Xu for the insightful discussions on SeqDPP which inspired this work.

{\small
\bibliographystyle{splncs}

}

\end{document}